\pdfoutput=1

\documentclass[11pt]{article}

\usepackage[final]{coling}

\usepackage{times}
\usepackage{latexsym}

\usepackage[T1]{fontenc}

\usepackage[utf8]{inputenc}

\usepackage{microtype}

\usepackage{inconsolata}

\usepackage{multirow}
\usepackage{xcolor}
\usepackage{bbm}
\usepackage[bottom]{footmisc}
\usepackage{float}
\usepackage{tikz}
\usepackage{xurl}
\usepackage{pgfplots}
\usepgfplotslibrary{groupplots}
\usepackage{graphics}
\usepackage{longtable}
\usepackage{caption}
\usepackage{hyperref}
\usepackage{subcaption}
\usepackage{algorithm}
\usepackage{algpseudocode}
\usepackage{mdframed}
\usepackage{booktabs}
\usepackage{graphicx}
\usepackage{amsmath}
\usepackage{balance}
\usepackage{xcolor}
\definecolor{CustomBlue}{RGB}{20, 81, 124}  
\definecolor{CustomLightBlue}{RGB}{47, 127, 193}  
\definecolor{CustomVeryLightBlue}{RGB}{231, 239, 250}  
\definecolor{CustomGreen}{RGB}{150, 195, 125}  
\definecolor{CustomYellow}{RGB}{243, 210, 102}  
\definecolor{CustomRed}{RGB}{216, 56, 58}  
\definecolor{CustomPink}{RGB}{247, 225, 237}  
\definecolor{CustomVeryLightPurple}{RGB}{248, 243, 249}  
\definecolor{CustomPurple}{RGB}{196, 151, 178}  
\definecolor{CustomGrayBlue}{RGB}{169, 184, 198}  
\title{Evaluating and Enhancing Large Language Models for Novelty Assessment in Scholarly Publications}
 
\author{Ethan Lin \thanks{These authors contributed equally to this work.} \\
  Santa Clara University\\
  Santa Clara, CA, USA \\
  \texttt{enlin@scu.edu} \\\And
  Zhiyuan Peng \footnotemark[1]\\
  Santa Clara University\\
  Santa Clara, CA, USA \\
  \texttt{zpeng@scu.edu} \\ \And
  Yi Fang\\
  Santa Clara University\\
  Santa Clara, CA, USA \\
  \texttt{yfang@scu.edu} \\}

\begin{document}
\maketitle
\begin{abstract}
 Recent studies have evaluated the creativity/novelty of large language models (LLMs) primarily from a semantic perspective, using benchmarks from cognitive science. However, accessing the novelty in scholarly publications is a largely unexplored area in evaluating LLMs. In this paper, we introduce a scholarly novelty benchmark (SchNovel \footnote{The SchNovel dataset and RAG-Novelty code are available at: https://github.com/ethannlin/SchNovel.}) to evaluate LLMs’ ability to assess novelty in scholarly papers. SchNovel consists of 15000 pairs of papers across six fields sampled from the arXiv dataset with publication dates spanning 2 to 10 years apart. In each pair, the more recently published paper is assumed to be more novel. Additionally, we propose RAG-Novelty, which simulates the review process taken by human reviewers by leveraging the retrieval of similar papers to assess novelty. Extensive experiments provide insights into the capabilities of different LLMs to assess novelty and demonstrate that RAG-Novelty outperforms recent baseline models.  
\end{abstract}

\section{Introduction}
Creativity, traditionally defined as the capability of producing innovative ideas, forming unique connections, and developing solutions that are both novel and effective \cite{runco2012standard}, was thought to be exclusive to humans. With the advent of large language models (LLMs) \cite{DBLP:journals/corr/abs-2303-08774, dubey2024llama}, however, this assumption has been challenged as LLMs have demonstrated remarkable proficiency in following human instructions to address a wide range of tasks, such as open-domain question answering, coding, solving math problems, and even utilizing external tools like search engines when necessary \cite{DBLP:conf/iclr/AsaiWWSH24}. In a recent study, \citet{si2024can} showed that LLMs can generate research ideas that are more novel than those produced by human experts. While their generative capabilities are impressive, they also raise concerns. For instance, some artists are uncomfortable with LLMs producing artwork that closely mimics established styles, prompting debates about whether LLMs can genuinely be considered creative.

Recent studies evaluating the generative creativity of LLMs have yielded inconsistent conclusions. \citet{orwig2024language} concluded that GPT-4 \cite{DBLP:journals/corr/abs-2303-08774} generates stories that are comparable to those written by humans in terms of creativity. Similarly, \citet{DBLP:journals/corr/abs-2405-13012} found that LLMs can even surpass humans in specific creative tasks, such as divergent association and creative writing. However, \citet{DBLP:conf/candc/AndersonSK24} argued that AI-based creativity support tools (CSTs) like ChatGPT are not yet well-suited to fostering truly original ideas, as they can lead to homogenization of human creativity. \citet{DBLP:conf/chi/ChakrabartyLAMW24} observed that LLM-generated stories pass the Torrance Test for Creative Writing (TTCW) tests 3 to 10 times less frequently than those written by professionals. Additionally, \citet{DBLP:journals/corr/abs-2309-12570} pointed out that LLMs often rely on clichés, produce text lacking nuance, and frequently resort to overly moralistic and predictable endings in stories. These discrepancies can be attributed to using different evaluation benchmarks and metrics, highlighting the lack of widely accepted standards for accessing LLM creativity. 

The evaluation benchmarks used in current studies are primarily derived from cognitive science, such as the Torrance Tests of Creative Thinking (TTCT) \cite{lissitz1985methodological}, Alternative Use Task (AUT) \cite{guilford1964some}, and the Runco Creativity Assessment Battery (rCAB) \cite{rCAB}. These benchmarks focus on assessing semantic creativity by tasks like generating responses to pictures or listing as many uses as possible for a common object. Corresponding metrics include fluency, flexibility, originality, and elaboration. However, these metrics primarily assess semantic novelty, which does not fully capture the the kind of novelty emphasized in scholarly research. Novelty in scholarly work is especially critical, as each paper undergoes rigorous peer review, particularly in high-prestige venues. Novel papers typically build upon existing research while introducing new ideas, methods, or insights, making novelty assessment heavily dependent on current and past trends in research.

While LLMs have shown great capability in generating text and mimicking human reasoning, their ability to assess novelty in scholarly publications remains largely unexamined. To address this gap, we present a scholarly novelty benchmark (SchNovel) to evaluate LLMs’ capability of assessing novelty in scholarly papers. Specifically, we leverage the arXiv dataset to create a collection of 15,000 paper pairs. In each pair, we assume that the more recently published paper is more novel. Papers are selected across six categories, with publication dates spaced by gaps ranging from 2 to 10 years between the paired papers. We evaluate various LLMs on their ability to assess novelty and report their accuracy.

To further improve novelty assessment, we propose RAG-Novelty, a retrieval-augmented generation method. This method assumes that more novel papers will retrieve more recently published works, enhancing the novelty prediction. Our extensive experiments demonstrate that RAG-Novelty outperforms recent baseline models in assessing novelty in scholarly papers. Our key contributions include:

\begin{itemize}
\item We release the first benchmark SchNovel specifically designed to evaluate LLMs’ capability in assessing novelty within scholarly publications.
\item We conduct comprehensive experiments to explore how variations in categories, starting years, and year gaps affect LLMs’ ability to assess paper novelty.
\item We propose a novel method RAG-Novelty to enhance LLMs’ performance in assessing paper novelty.
\end{itemize}

\section{Related Work}
\subsection{Existing Benchmarks} \label{sec: benchmarks}
TTCT \cite{lissitz1985methodological} is a commercially protected assessment tool consisting of six tasks: 1) asking a question about a picture; 2) guessing the cause of the action depicted in the image; 3) predicting the consequences of the action described in the image; 4) improving a product described in 2-3 sentences in the most interesting and unusual way; 5) suggesting interesting and unconventional uses for a given item; and 6) imagining what would happen if an improbable situation were to occur. Both AUT \cite{guilford1964some} and rCAB \cite{rCAB} ask participants to generate as many uses as possible for a common object. The Remote Associates Test (RAT) \cite{mednick1968remote} presents participants with three seemingly unrelated words and asks them to find a fourth word that connects all three. The Consensual Assessment Technique (CAT) \cite{amabile1982social} evaluates creative products, such as stories, poetry, dramatic performances, and musical compositions, using a panel of domain experts. The Wallach-Kogan Creativity Tests (WCT) \cite{brody1966modes} consist of the AUT, Instances Test, and Similarities Test. The scholarly Creativity Test (SCT) \cite{hu2002scientific} measures scholarly creativity and process skills. The Divergent Association Task (DAT) \cite{olson2021naming} asks participants to name unrelated nouns and calculates the pairwise semantic distance between them. However, all these existing cognitive science benchmarks are not suited for evaluating LLMs’ capability to assess novelty in scholarly publications, a gap our proposed benchmark addresses.

\subsection{Creativity/Novelty Assessment}
Traditional general novelty assessment methods use pre-defined metrics like the similarity to existing methods \cite{just2024ai} and the diversity of references \cite{shibayama2021measuring} to score the novelty of a method or scholarly paper. To assess LLMs’ capability of generating or assessing creativity and novelty, current studies employ different prompt strategies to interact with LLMs and collect responses for evaluation. \citet{guzik2023originality} utilized a basic prompt to evaluate GPT-4 on the TTCT benchmark. \citet{DBLP:journals/corr/abs-2405-06715} applied associative thinking \cite{mednick1962associative} in prompts designed for specific tasks like product design and marketing. \citet{DBLP:journals/corr/abs-2401-12491} analyzed LLMs’ responses to an expanded TTCT benchmark, applying diverse prompts, including basic prompts, instructive prompts, post-instructive prompts, and Chain of Thought (CoT) prompts. \citet{DBLP:conf/icccrea/StevensonSBGM22} demonstrates that defining the role of LLMs as “scientist” can improve performance. \citet{summers2023brainstorm} improves the basic prompt method used in \cite{DBLP:conf/icccrea/StevensonSBGM22} by using multi-step reasoning to enhance GPT-3’s performance on AUT. Similar to the multi-round interaction framework utilized in LLM Debate \cite{DBLP:conf/icml/Du00TM24}, LLM Discussion \cite{DBLP:journals/corr/abs-2405-06373} develops a role-play enhanced LLM discussion framework to augment ChatGPT’s performance on the WCT and SCT benchmarks.
Unlike existing prompting methods, our proposed RAG-Novelty improves the LLM's performance by retrieving similar papers assuming that novel papers should retrieve the latest publications.

\subsection{LLM Performance Evaluation}
Most existing studies \cite{summers2023brainstorm, DBLP:conf/icccrea/StevensonSBGM22, guzik2023originality, mednick1962associative} evaluate LLM performance on benchmarks (Section \ref{sec: benchmarks}) using human assessments. For example, \citet{guzik2023originality} evaluated LLM responses to the TTCT, which were scored by Scholastic Testing Services (STS). Other studies rely on LLMs to score responses from another LLM. \citet{DBLP:journals/corr/abs-2401-12491} used a more powerful GPT-4 to evaluate the performance of smaller LLMs, while \citet{DBLP:journals/corr/abs-2405-06373} utilized ChatGPT to assess responses generated by GPT-4. Additionally, \citet{DBLP:journals/corr/abs-2405-06373} compared LLM-generated scores with human evaluations, finding that LLM evaluations correlated more closely with the average human score. Both \citet{luchini2023automatic} and \cite{organisciak2023beyond} fine-tuned models on human-scored data to evaluate LLM responses. Since our benchmark provides ground-truth binary labels, evaluation is straightforward.

\section{Scholarly Novelty Benchmark}
Unlike the semantic novelty evaluated by the benchmarks from cognitive science (Section \ref{sec: benchmarks}), novelty in scholarly publications refers to introducing new ideas, methods, or discoveries that have previously not been explored or established in the literature. Evaluating novelty is fundamentally an exercise in understanding the relationship between ideas across time rather than simply assessing new ideas or techniques. This understanding is crucial in determining the contribution of a research paper. The assumption can be made that later works are more novel than prior works, as they typically introduce new ideas and methodologies in the current research climate. In this paper, we apply this assumption to establish ground truth values for our created benchmark SchNovel.

\subsection{Dataset Collection and Structure}

The arXiv dataset\footnote{Available at \url{https://www.kaggle.com/datasets/Cornell-University/arxiv}} comprises approximately 2.5 million articles, with the earliest dating back to 1986. All articles are categorized into eight distinct fields\footnote{See the full taxonomy at \url{https://arxiv.org/category_taxonomy}} each of which has some sub-fields. We picked up six out of eight fields: Computer Science (cs), Mathematics (math), Physics (physics), Quantitative Biology (q-bio), Quantitative Finance (q-fin), and Statistics (stat), as we did not collect enough papers in other fields. Figure \ref{fig: num-per-year-arxiv} in Appendix \ref{ap: stat-arxiv} shows the number of papers published each year for each field. To assess the ability of LLMs to assess the novelty of research papers, we sampled a subset of articles from each field, denoted as dataset $D=\{(f, g, s, x, y, label)_{i}\}_{i=1}^{N}$ where $N=15000$, following the procedure outlined in Algorithm \ref{alg: sampling} in Appendix \ref{ap: scinovel}, where $f$ represent the field, $x$ and $y$ represent the paper ids, $s$ represents the year in which paper $x$ was published, $g$ represents the number of years paper $y$ was published before paper $x$ and $label$ equals to paper $x$ as we assume in the same field, later published paper is more novel. 

\subsection{Tasks and Evaluation Metrics}
We define the task as assessing which paper is more novel when given a pair of papers. Specifically, for each tuple $(f, g, s, x, y, label)_{i}$, the selected LLM is provided with the title, abstract, and optional metadata for each paper—information typically available to a reviewer. However, unlike a full review, the model does not have access to the full text, making the task more challenging. While the abstract offers a strong indication of a paper’s content and key findings, important details may be missed. By limiting the context to the abstract and metadata, we also improve efficiency in terms of token consumption and cost. We will discuss the potential limitations of this approach in Section \ref{sec: limitations}. Various comparison methods, such as point-wise and pair-wise, can be employed, and we evaluate performance based on accuracy.
\section{RAG-Novelty}

\begin{figure} [t] 
  \centering
  \includegraphics[width=0.9\linewidth]{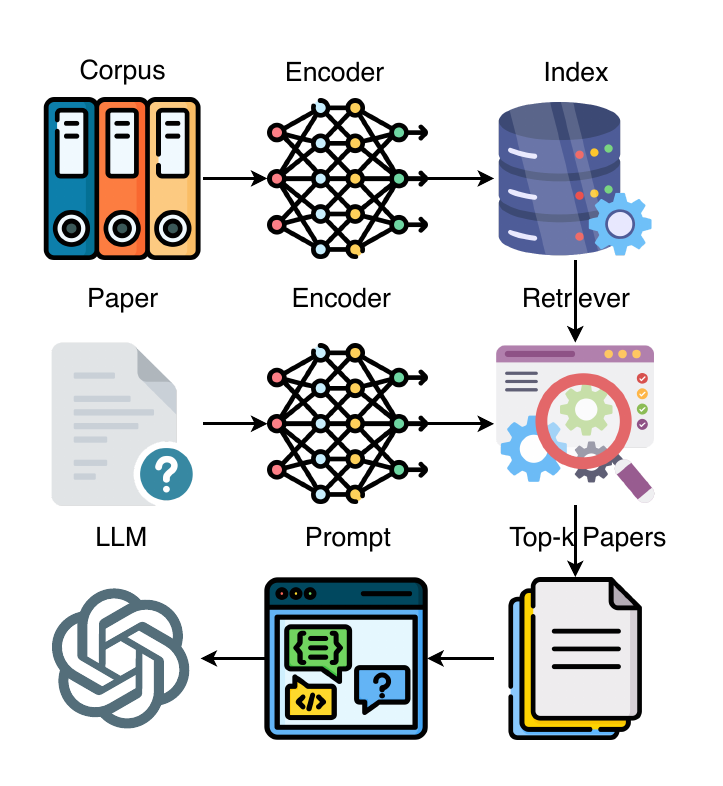}
  \caption{The overview of RAG-Novelty}
  \label{fig: main_diagram}
\end{figure}

Assessing the novelty in scholarly papers requires the model to have a good understanding of past and present works to accurately judge whether a paper is novel in the current research climate. However, once trained, LLMs are frozen in time, meaning that they are no longer updated with the latest information, so they lack this understanding of the field's current state. Inspired by RAG, we propose a novel method, RAG-Novelty, to further improve LLMs' capability to assess novelty in our benchmark. As shown in Figure \ref{fig: main_diagram}, apart from the information, like abstract, that can be utilized for a paper, we apply the paper abstract as a query to retrieve top-K papers from the already built index, and then we create a prompt based on the query paper and the retrieved papers to ask LLM score the novelty of query paper from 0 to 10. 

\subsection{Indexing and Retriever}
To assess the novelty of a paper with the information provided by our SchNovel, such as title, abstract, and other metadata excluding the whole paper, an expert human reviewer in the same field may accurately score the novelty, a junior human reviewer, however, is likely not confident of scoring the novelty directly and instead, he/she will first review some similar papers and then assess the novelty. To mimic the review process taken by a human reviewer, we randomly sampled 500 papers from all years from 2000 to 2023, yielding 12000 papers for each field. Then, the abstracts of these papers are encoded into embeddings using OpenAI's \textit{text-embedding-3-small\footnote{\url{https://platform.openai.com/docs/guides/embeddings}}} model. The retrieve is the exact search method based on cosine similarity, as the number of candidates is very small. Our method can also handle huge candidate corpus by building an approximate nearest neighbor searching index using faiss \cite{douze2024faiss, johnson2019billion}. 

When a human reviewer conducts a literature search, it is naturally impossible to retrieve papers published after the query paper’s publication date. To simulate this realistic constraint in our evaluation, we filtered out any papers published after the query paper and retrieved the top-k relevant papers from those published prior to or on the same date. However, in the context of pairwise comparisons, where we are assessing the novelty between two papers with different publication dates, it is reasonable to retrieve papers up to the publication date of the more recent paper. This approach mirrors a realistic scenario in which novelty is judged relative to the latest available knowledge at the time of publication. By implementing this strategy, we ensure that the novelty assessment remains fair and contextually appropriate, avoiding any temporal bias while maintaining the integrity of the comparison.

\subsection{Prompt}
We first compared the zero-shot, two-shot, and self-reflection prompts and found that the self-reflection prompt performed the best (Section xx). So, for RAG-Novelty, we built the prompt, shown in Appendix \ref{ap: prompt-rag-novelty}, based on the self-reflection prompt, shown in Appendix \ref{ap: prompt-self-reflection}, by incorporating the information of the retrieved papers. Specifically, we added a ``Contextual Data Analysis'' instruction that assumes that the more latest papers are retrieved, the more novel this query paper is:

 \textit{\small Average the published dates of the retrieved documents. Use this average date as additional context for your evaluation. Consider that
papers with an average date that is later or more recent
in time are generally more novel.}

\section{Experimental Setup}

\subsection{Baseline Methods} \label{sec: baselines}
\textbf{Zero-Shot} as shown in Appendix \ref{ap: zero-shot-prompt}, involves providing the model with two research papers’ titles, abstracts, and four-step instructions, guiding the LLM to leverage its internal knowledge to make an informed decision. We also conducted the pointwise comparison by revising the zero-shot prompt to instruct the LLM score on the novelty of each paper first and then compare which one is more novel.\\
\textbf{Two-Shot} We randomly sampled two example paper pairs and added them to the zero-shot prompt.\\
\textbf{Chain of Thought (CoT)} \cite{wei2023chainofthoughtpromptingelicitsreasoning} elicits reasoning within models by giving
the model time to ``think''. We achieved CoT by adding instructions to Zero-Shot guiding LLMs to provide demonstrations.\\
\textbf{Self-Reflection} \cite{renze2024selfreflectionllmagentseffects} has shown several strides at improving LLMs logical fallacies by prompting the model to reflect on its incorrect solutions. We adopted this strategy to design a prompt which is shown in Appendix \ref{ap: prompt-self-reflection}. \\
\textbf{Self-Consistency} \cite{DBLP:conf/iclr/0002WSLCNCZ23} assumes that ground truth answers can be achieved through different reasoning paths. We followed the original paper to sample 10 generated sequences and voted majority.\\
\textbf{LLM Discussion} \cite{DBLP:journals/corr/abs-2405-06373} assigns LLMs with different roles and lets them discuss with each other before making the final decision. We adopted LLM Discussion to simulate the review process taken by human reviewers. Specifically, we assume the papers are submitted to a conference to be reviewed, and we designed four roles: (a) a professor; (b) a PhD student; (c) an editor of a prestigious journal; (d) the chair of the conference where the professor, PhD student, and editor are all reviewers and they have two round discussions and the chair make the final decision. The prompt is shown in Appendix \ref{ap: LLM Discussion}.

\subsection{LLM Configuration}
We adopted the default settings of API \footnote{\url{https://platform.openai.com/docs/guides/chat-completions}} for Zero-Shot, Two-Shot, CoT and RAG-Novelty. We followed the Self-Consistency to adopt the temperature as 0.7 and set the number of reasoning paths as 10. For LLM discussion, we limit the max tokens to 200 to avoid overwhelming the model with long inputs in subsequent rounds of discussion. For Self-consistency, we limit the max tokens so that the response is concise, as long reasoning for this task is unnecessary because we're looking for consistency rather than depth. In both cases, we prompt the model to limit its output to 150 tokens to ensure that it's response fits within the 200 token limit.

\subsection{Research Questions}
This study aims to address several key questions regarding the performance of LLMs on the SicNovel benchmark. 

\begin{itemize}
\item \textbf{R1:} Which comparison approach yields better results: pointwise or pairwise?
\item \textbf{R2:} How do different LLMs perform in assessing the novelty of research papers?
\item \textbf{R3} How does the category of the research paper affect the performance of LLMs?
\item \textbf{R4:} How does the publication start year influence the performance of LLMs?
\item \textbf{R5:} What impact does the gap between the publication years of research papers have on LLMs' performance?
\item \textbf{R6:} What are the effects of other metadata attributes on LLMs' performance?
\item \textbf{R7:} Can RAG-Novelty outperform recent baselines?
\end{itemize}

\begin{table}[t!]
\centering
\resizebox{\columnwidth}{!}{%
\begin{tabular}{c|cccccc}
\toprule
Method & cs & math & physics & qbio & qf\-in & stat\\
\midrule
Zero-Shot & 0.64 & 0.55 & 0.57 & 0.54 & 0.55 & 0.63 \\
Two-Shot & 0.62 & 0.55 & 0.57 & 0.54 & 0.55 & 0.60 \\
CoT & 0.63 & 0.56 & 0.57 & 0.54 & 0.56 & 0.62 \\
Self-Reflection & 0.65 & 0.56 & 0.58 & 0.56 & 0.57 & 0.63 \\
LLM Discussion & 0.60 & 0.55 & 0.56 & 0.53 & 0.50 & 0.58 \\
Self-Consistency & \underline{0.66} & \underline{0.57} & \underline{0.59} & \underline{0.58} & \underline{0.60} & \underline{0.64} \\
RAG-Novelty & $\dagger0.72^\star$ & $0.58^\star$ & $\dagger0.62^\star$ & $\dagger0.65^\star$ & $\dagger0.73^\star$ & $\dagger0.68^\star$ \\
\bottomrule
\end{tabular}}
\caption{RAG-Novelty vs. Baselines on SciEval with GPT-4o-mini. Averaged accuracy is reported. $\dagger$ denotes statistically significant enhancements over the second-best result, with p-values < 0.05, as determined by the McNemar test. The best results across different methods are denoted with symbol $\star$. The second-best results across different methods are underlined.}
\label{tab:RAG vs Baselines}
\end{table}

\section{Experimental Results}
\subsection{RAG-Novelty vs. Baseline Models (R7)}
In this experiment, we evaluate the performance of RAG-Novelty against baseline methods. All methods use GPT-4o-mini, and the accuracy is averaged across different start years $s$ and year gaps $g$. Pairwise comparison is applied to all methods, and we account for position bias by swapping the order of the two papers in the comparisons.

Two-Shot does not improve upon Zero-Shot as it typically does in other tasks. We attribute this to the complexity of the novelty assessment task, which requires deeper contextual understanding and comparison between papers—something that randomly selected examples may not effectively convey. Through iterative prompt refinement, Self-Reflection outperforms CoT in all fields except mathematics. LLM Discussion methods perform the worst, failing to even surpass Zero-Shot. Self-Consistency achieves the best results among baseline methods, demonstrating that obtaining answers through different reasoning paths helps improve performance. Our RAG-Novelty achieves the highest results overall, significantly outperforming the second-best method, except in the mathematics field. Across all methods, the improvement in mathematics is limited, possibly due to the slower progression of the field, the prevalence of symbols that LLMs struggle to interpret, or a lack of sufficient mathematical content in the training data compared to other fields.

\subsection{Pointwise vs Pairwise (R1)}

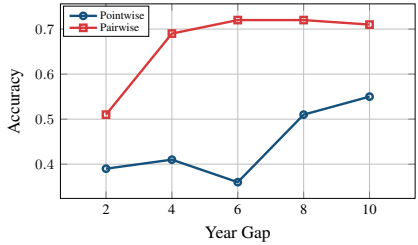
\begin{figure}[t!]
    \centering
    \begin{tikzpicture}[scale=0.6]
        \begin{axis}[
            height=5.88cm, 
            width=9.36cm,
            ylabel={Accuracy}, 
            xlabel={Year Gap},
            symbolic x coords={2,4,6,8,10}, 
            xtick=data, 
            enlarge x limits={abs=1cm}, 
            label style={font=\small}, 
            tick label style={font=\small},
            xlabel near ticks, 
            ylabel near ticks, 
            legend style={nodes={scale=0.65, transform shape}, at={(0.25,0.99)}, 
            /tikz/every odd column/.style={yshift=2pt},
            /tikz/nodes={text width=40pt, text depth=, anchor=base}},
            grid=major  
            ]
            \addplot[mark=o, line width=1.5pt, color=CustomBlue] coordinates {(2,0.39) (4,0.41) (6,0.36) (8,0.51) (10,0.55)};
            
            \addplot[mark=square, line width=1.5pt, color=CustomRed] coordinates {(2,0.51) (4,0.69) (6,0.72) (8,0.72) (10,0.71)};
            
            \legend{Pointwise, Pairwise}
        \end{axis}
    \end{tikzpicture}
    \caption{Pointwise vs. pairwise. The metrics above were obtained in the cs field with the start year $s=2023$ and GPT-4o-mini.}
    \label{fig: pointwise_pairwise_chart}
\end{figure}

As mentioned in Section \ref{sec: baselines}, we revised the pairwise Zero-Shot prompt (Appendix \ref{ap: zero-shot-prompt}) to a pointwise one. We compared the two methods by evaluating them in the cs field with the start year 2023, crossing different year gaps. As shown in Figure \ref{fig: pointwise_pairwise_chart}, pairwise is consistently much better than pointwise across different year gaps. This significant difference highlights the importance of context. As with human evaluations, providing relevant context or reference points is crucial for accurate assessments \cite{10.1145/3477495.3531991}, allowing reviewers to consider the broad implications of a paper within the current research landscape. Pairwise comparisons align with this process, simplifying the task of considering the relative merits of two papers side-by-side rather than evaluating each one in isolation. Thus, pairwise comparisons are used in the rest of the following experiments.
\subsection{The Impact of Different Fields (R3)}

Different research domains require varying expertise to evaluate novelty. This section examines how different fields influence LLM performance in novelty evaluation.

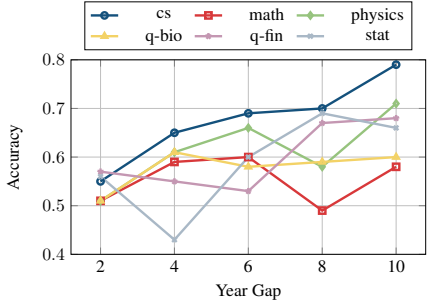
\begin{figure}[t]
    \centering
    \begin{tikzpicture}[scale=0.6]
        \begin{axis}[
            height=5.88cm, 
            width=9.36cm,
            xlabel={Year Gap},
            ylabel={Accuracy},
            symbolic x coords={2, 4, 6, 8, 10},
            xtick=data,
            ymin=0.4, ymax=0.8,
            grid=major,
            legend style={
                at={(0.5,1.05)},  
                anchor=south,    
                legend columns=3, 
                column sep=1ex, 
                /tikz/every odd column/.append style={column sep=0.5cm} 
            },
            mark options={solid},
            ]
            
            \addplot[mark=o, color=CustomBlue, line width=1.5pt] coordinates {
                (2,0.55) (4,0.65) (6,0.69) (8,0.70) (10,0.79)
            };
            \addplot[mark=square, color=CustomRed, line width=1.5pt] coordinates {
                (2,0.51) (4,0.59) (6,0.60) (8,0.49) (10,0.58)
            };
            \addplot[mark=diamond, color=CustomGreen, line width=1.5pt] coordinates {
                (2,0.51) (4,0.61) (6,0.66) (8,0.58) (10,0.71)
            };
            \addplot[mark=triangle, color=CustomYellow, line width=1.5pt] coordinates {
                (2,0.51) (4,0.61) (6,0.58) (8,0.59) (10,0.60)
            };
            \addplot[mark=star, color=CustomPurple, line width=1.5pt] coordinates {
                (2,0.57) (4,0.55) (6,0.53) (8,0.67) (10,0.68)
            };
            \addplot[mark=x, color=CustomGrayBlue, line width=1.5pt] coordinates {
                (2,0.56) (4,0.43) (6,0.60) (8,0.69) (10,0.66)
            };

            \legend{cs, math, physics, q-bio, q-fin, stat}
        \end{axis}
    \end{tikzpicture}
    \caption{Comparison of fields. The metrics above were obtained using Self-Reflection in cs field with the start year $s=2023$ with GPT-4o-mini.}
    \label{fig: fields}
\end{figure}

In Figure \ref{fig: fields}, the cs category shows the highest accuracy across most year gaps (starting in 2023), likely due to the availability of data and well-defined evaluation metrics. In contrast, math and physics show lower accuracy, likely due to domain-specific challenges such as complex notation in mathematics and theoretical frameworks in physics.

One explanation is the lack of domain knowledge in ChatGPT’s training data, which, being sourced from the internet, may not adequately cover specialized fields. Research has shown that LLMs exhibit biases in various prompts and tasks \cite{cheng-etal-2023-marked, stranisci-etal-2023-wikibio}, suggesting potential categorical biases in lesser-known or slower-growing domains. This has significant implications for using AI tools in academia and industry, particularly in automated scoring or ranking systems, where such biases could perpetuate inequalities.

\subsection{The Impact of Different Start Years and Year Gaps (R4 \& R5)}

\begin{figure}[t]
    \centering
    \begin{tikzpicture}[scale=0.6]
        \begin{axis}[
            height=5.88cm, 
            width=9.36cm,
            xlabel={Year Gap},
            ylabel={Accuracy},
            symbolic x coords={2, 4, 6, 8, 10},
            xtick=data,
            ymin=0.4, ymax=0.8,
            grid=major,
            legend style={
                at={(0.5,1.05)},  
                anchor=south,    
                legend columns=3, 
                column sep=1ex, 
                /tikz/every odd column/.append style={column sep=0.5cm} 
            },
            mark options={solid},
            ]
            
            \addplot[mark=o, color=CustomBlue, line width=1.5pt] coordinates {
                (2,0.54) (4,0.64) (6,0.69) (8,0.65) (10,0.72)
            };
            \addplot[mark=square, color=CustomRed, line width=1.5pt] coordinates {
                (2,0.49) (4,0.61) (6,0.65) (8,0.71) (10,0.76)
            };
            \addplot[mark=diamond, color=CustomGreen, line width=1.5pt] coordinates {
                (2,0.59) (4,0.57) (6,0.60) (8,0.72) (10,0.71)
            };
            \addplot[mark=triangle, color=CustomYellow, line width=1.5pt] coordinates {
                (2,0.585) (4,0.57) (6,0.70) (8,0.71) (10,0.72)
            };
            \addplot[mark=star, color=CustomPurple, line width=1.5pt] coordinates {
                (2,0.55) (4,0.65) (6,0.69) (8,0.70) (10,0.79)
            };

            \legend{2019, 2020, 2021, 2022, 2023}
        \end{axis}
    \end{tikzpicture}
    \caption{Comparison of Start Years. The metrics above were obtained using Self-Reflection in the cs field with GPT-4o-mini.}
    \label{fig: start-gap}
\end{figure}
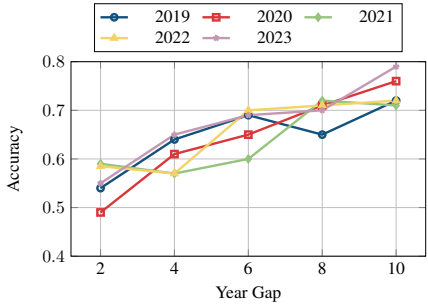

To better understand how different start years affect the performance of LLMs in evaluating novelty, we investigated the model’s results for five distinct start years. As shown in Figure \ref{fig: start-gap}, the model’s results for all five start years were relatively consistent across different year gaps. This suggests that the model’s ability to evaluate novelty between two papers is more dependent on the year gap between them than the specific publication years.

For example, evaluating two papers with a 10-year gap from 2009 to 2019 should be equivalent in difficulty to evaluating two papers with a 10-year gap from 2013 to 2023. Regardless of the boundary years within those ranges (i.e., considering papers published at specific points like 2009 and 2019, versus 2013 and 2023), it's the decade-long gap between the papers' publication times that makes it easier for the model to make such a binary evaluation.

\begin{table*}[t!]  
\centering
\resizebox{\textwidth}{!}{  
\begin{tabular}{c|ccc|ccc|ccc|ccc|ccc|ccc}
\toprule
\textbf{Year Gap} & \multicolumn{3}{c|}{\textbf{ChatGPT4o-mini}} & \multicolumn{3}{c|}{\textbf{ChatGPT3.5}} & \multicolumn{3}{c|}{\textbf{LLaMA 3.1-8b}} & \multicolumn{3}{c|}{\textbf{Mistral-7b}} & \multicolumn{3}{c|}{\textbf{Gemma-2-9b}} \\
\cmidrule(lr){2-4} \cmidrule(lr){5-7} \cmidrule(lr){8-10} \cmidrule(lr){11-13} \cmidrule(lr){14-16} \cmidrule(lr){17-19}
 & Asc Yr & Desc Yr & Acc. & Asc Yr & Desc Yr & Acc. & Asc Yr & Desc Yr & Acc. & Asc Yr & Desc Yr & Acc. & Asc Yr & Desc Yr & Acc. \\
\midrule
2 & 0.44 & 0.66 & 0.55 & 0.46 & 0.62 & 0.54 & 0.03 & 0.98 & 0.51 & 1.00 & 0.00 & 0.50 & 0.66 & 0.38 & 0.52\\
4 & 0.58 & 0.72 & 0.65 & 0.58 & 0.57 & 0.58 & 0.02 & 0.97 & 0.50 & 1.00 & 0.00 & 0.50 & 0.70 & 0.48 & 0.59 \\
6 & 0.63 & 0.75 & 0.69 & 0.67 & 0.60 & 0.64 & 0.01 & 0.99 & 0.50 & 1.00 & 0.01 & 0.51 & 0.69 & 0.41 & 0.55 \\
8 & 0.63 & 0.77 & 0.70 & 0.63 & 0.68 & 0.66 & 0.01 & 0.99 & 0.50 & 0.99 & 0.00 & 0.50 & 0.76 & 0.46 & 0.61 \\
10 & 0.79 & 0.78 & 0.79 & 0.67 & 0.71 & 0.69 & 0.05 & 0.97 & 0.51 & 0.99 & 0.01 & 0.50 & 0.80 & 0.43 & 0.62 \\
\hline
Average & 0.61 & 0.74 & 0.68 & 0.60 & 0.64 & 0.62 & 0.02 & 0.98 & 0.50 & 0.996 & 0.004 & 0.50 & 0.72 & 0.43 & 0.58 \\
\bottomrule
\end{tabular}
}
\caption{Comparision of different LLMs. The metrics above were obtained using Self-Reflection in the cs field with the start year $s=2023$. ``Asc Yr'' indicates that the older paper is presented first in the prompt, while ``Desc Yr`` means the newer paper is presented first.}
\label{tab:llms}
\end{table*}

\subsection{The Impact of Different LLMs (R2)}
All LLMs can vary significantly depending on their training data and model architecture. With various different models available, it is essential to understand how they perform when assessing the originality of ideas presented in research papers. In this section, we examine the impact of using different LLMs on evaluating novelty.

Our findings in Table~\ref{tab:llms} reveal significant disparities in the performance across different LLMs. GPT-4o-mini, GPT-3.5, and Gemma 2 performed more in line with expectations, achieving a more balanced distribution of predictions throughout all year gaps. Notably, GPT-4o-mini outperformed all other models, demonstrating a substantial advantage over the smaller models like LLaMA 3.1-8b, Mistral 7b, and Gemma 2-9b.


Despite such success, even ChatGPT 4o-mini and ChatGPT 3.5 exhibit position bias, where the order of papers in the prompt affects their decision-making instead of content alone. This bias is magnified in smaller models, which lack extensive training compared to larger models. For example, Mistral 7b is heavily biased towards the second paper, which comes last in the prompt. This aligns with known issues regarding LLMs' performance being best when relevant information appears towards the beginning or end of the prompt \cite{lostinthemiddle, ir+bias}.

In contrast, LLaMA 3.1-8b exhibits a different bias – favoring the first paper that appears towards the middle of the prompt. According to \citet{dubey2024llama3herdmodels}, the LLaMA 3.1 models excel at "needle-in-the-haystack" tasks, where one needs to find specific information in large amounts of text \cite{kamradt2023llmtest}, ultimately fixing the issues described in \citet{lostinthemiddle}. This is similar to skimming, which is efficient for finding specific information but may not facilitate deep understanding. Likewise, LLaMA 3.1-8b excels at retrieving specific information from anywhere in a context but this skillset is not ideal for evaluating novelty between two papers.

\begin{table*}[t!]  
\centering
\fontsize{6.5}{8}\selectfont  
\begin{tabular}{c|ccc|ccc|ccc|ccc|ccc}
\toprule
\textbf{Year Gap} & \multicolumn{3}{c|}{\textbf{Zero-Shot}} & \multicolumn{3}{c|}{\textbf{Self-Reflection}} & \multicolumn{3}{c|}{\textbf{Self-Reflection w/ tldr}} & \multicolumn{3}{c|}{\textbf{Self-Reflection 2 w/ author}} \\
\cmidrule(lr){2-4} \cmidrule(lr){5-7} \cmidrule(lr){8-10} \cmidrule(lr){11-13} \cmidrule(lr){14-16}
 & Asc Yr & Desc Yr & Acc. & Asc Yr & Desc Yr & Acc. & Asc Yr & Desc Yr & Acc. & Asc Yr & Desc Yr & Acc.\\
\midrule
2  & 0.41 & 0.60 & 0.51 & 0.44 & 0.66 & 0.55 & \textbf{0.53} & \textbf{0.51} & 0.52 & \textbf{0.57} & \textbf{0.55} & 0.56\\
4  & 0.63 & 0.74 & 0.69 & 0.58 & 0.72 & 0.65 & \textbf{0.64} & \textbf{0.64} & 0.64 & 0.67 & 0.62 & 0.65\\
6  & 0.64 & 0.79 & 0.72 & 0.63 & 0.75 & 0.69 & \textbf{0.66} & \textbf{0.69} & 0.68 & 0.75 & 0.62 & 0.69\\
8  & 0.66 & 0.77 & 0.72 & 0.63 & 0.77 & 0.70 & 0.69 & 0.61 & 0.65 & \textbf{0.68} & \textbf{0.67} & 0.68\\
10 & 0.64 & 0.78 & 0.71 & 0.78 & 0.79 & 0.79 & \textbf{0.76} & \textbf{0.76} & 0.76 & 0.80 & 0.75 & 0.78\\
\hline
Average & 0.60 & 0.74 & 0.67 & 0.61 & 0.74 & 0.68 & 0.66 & 0.64 & 0.65 & 0.69 & 0.64 & 0.67 \\
\bottomrule
\end{tabular}
\caption{The impact of metadata. The metrics above were obtained using Self-Reflection in the cs field with the start year $s=2023$ and GPT-4o-mini. ``Asc Yr'' indicates that the older paper is presented first in the prompt, while ``Desc Yr`` means the newer paper is presented first.}
\label{tab: metadata}
\end{table*}

\subsection{The Impact of Metadata (R6)}
Previously, our experiments evaluated novelty based solely on a paper’s title and abstract. However, human evaluations often take into account various metadata that can subtly influence reviewers' decisions. This metadata-induced bias has significant implications for research evaluations and highlights the need for more anonymous reviewal processes leading to solutions such as double-blind reviewal processes. In this section, we explore how metadata affects LLMs' ability to evaluate novelty. A pairwise comparison was applied for all the experiments in this section, and we accounted for position bias by swapping the order of the two papers in the comparisons.

\subsubsection{Adding a TLDR Summary}
We utilized the SciTLDR model \cite{cachola2020tldrextremesummarizationscientific} from the Semantic Scholar API \cite{Kinney2023TheSS} to generate TLDRs for our dataset, expecting this additional information to enhance accuracy through the addition of context to help the model generalize and better understand the paper. As shown in Table \ref{tab: metadata}, adding TLDRs decreases the accuracy across all year gaps. Nevertheless, incorporating such data did mitigate position bias, as evidenced by the almost negligible difference between ascending and descending year accuracies across nearly all year gaps.

\subsubsection{Adding Author}
We then added the author to the prompt, expecting that this additional information would not affect the model performance as the authors should not influence the novelty assessment. To our surprise, adding such information did help mitigate some of the position bias, as seen in the bold results in Table \ref{tab: metadata}, but overall decreased the performance slightly. 

\subsubsection{Adding Affiliation}
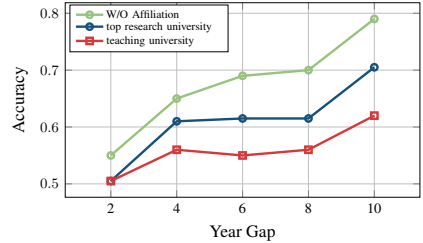
\begin{figure}[t!]
    \centering
    \begin{tikzpicture}[scale=0.6]
        \begin{axis}[
            height=5.88cm, 
            width=9.36cm,
            ylabel={Accuracy}, 
            xlabel={Year Gap},
            symbolic x coords={2,4,6,8,10}, 
            xtick=data, 
            enlarge x limits={abs=1cm}, 
            label style={font=\small}, 
            tick label style={font=\small},
            xlabel near ticks, 
            ylabel near ticks, 
            legend style={nodes={scale=0.65, transform shape}, at={(0.46,0.99)}, 
            /tikz/every odd column/.style={yshift=2pt},
            /tikz/nodes={text width=110pt, text depth=, anchor=base}},
            grid=major  
            ]
            \addplot[mark=o, line width=1.5pt, color=CustomGreen] coordinates {(2,0.55) (4,0.65) (6,0.69) (8,0.70) (10,0.79)};
            
            \addplot[mark=o, line width=1.5pt, color=CustomBlue] coordinates {(2,0.505) (4,0.61) (6,0.615) (8,0.615) (10,0.705)};
            
            \addplot[mark=square, line width=1.5pt, color=CustomRed] coordinates {(2,0.505) (4,0.56) (6,0.55) (8,0.56) (10,0.62)};
            
            \legend{W/O Affiliation, top research university, teaching university}
        \end{axis}
    \end{tikzpicture}
    \caption{Comparison of different organizations. The metrics above were obtained using Self-Reflection in the cs field with start year $s=2023$ and GPT-4o-mini.}
    \label{fig: org_bias}
\end{figure}

We selected two universities, one of which is a top research university and the other a teaching university, to study whether affiliation bias exists in LLMs’ assessment of novelty.\footnote{The real names of the universities are not used to ensure objectivity and to avoid any unintended bias or implications.} Specifically, we first assigned the top research university as the affiliation of the more recently published paper and the teaching university to the earlier published paper, with the results shown in blue. Then, we swapped the affiliations, and the results are shown in red. As illustrated in Figure \ref{fig: org_bias}, the top research university starts with similar accuracy to the teaching university at a year gap of $g=2$, but as the year gap increases, the top research university consistently outperforms the teaching university. This suggests that affiliation bias exists in LLMs’ novelty assessments, with a tendency to “trust” papers from top research universities. However, even though we observed LLMs' preference for choosing the top research university, the top research university experiments are outdone by W/O affiliation. This unexpected result raises questions about how LLMs process affiliation information, which warrants further investigation to better understand and mitigate such biases.

\section{Conclusion and Future Work}
To evaluate LLMs’ ability to assess novelty in scholarly publications, we introduce SchNovel, a benchmark consisting of 15,000 pairs of papers across six fields. We conducted extensive experiments to understand how various factors influence LLM performance on SchNovel. To enhance LLMs’ capability in assessing novelty, we propose RAG-Novelty, which significantly outperforms strong baseline models in comprehensive experiments. For future work, we plan to expand SchNovel by including more papers and covering additional fields to evaluate LLMs on a larger scale. Another promising direction is investigating which part of a paper best represents the whole for novelty assessment by LLMs. Additionally, studying how LLMs process affiliation and addressing biases in novelty evaluation, such as position and affiliation bias, is an important area for further research.


\clearpage
\newpage
\section{Limitations} \label{sec: limitations}
Our study evaluates an LLM’s ability to assess novelty using a research paper’s title, abstract, and metadata. While the abstract provides a strong indication of a paper’s content and key findings, it may not fully capture the novelty of the research compared to the complete text. Abstracts often summarize the main ideas but may omit important technical details. Although this approach streamlines the evaluation process, it could occasionally limit the depth of the novelty assessment due to the absence of a more comprehensive context.
\bibliography{custom}

\appendix

\section{Appendix}

\subsection{Statistics of arXiv} \label{ap: stat-arxiv}


\begin{figure}[H]
    \centering
    \resizebox{0.8\linewidth}{!}{
\begin{tikzpicture}

\definecolor{crimson2143940}{RGB}{214,39,40}
\definecolor{darkgray176}{RGB}{176,176,176}
\definecolor{darkorange25512714}{RGB}{255,127,14}
\definecolor{forestgreen4416044}{RGB}{44,160,44}
\definecolor{lightgray204}{RGB}{204,204,204}
\definecolor{mediumpurple148103189}{RGB}{148,103,189}
\definecolor{sienna1408675}{RGB}{140,86,75}
\definecolor{steelblue31119180}{RGB}{31,119,180}

\begin{axis}[
legend cell align={left},
legend style={
  fill opacity=0.8,
  draw opacity=1,
  text opacity=1,
  at={(0.03,0.97)},
  anchor=north west,
  draw=lightgray204
},
tick align=outside,
tick pos=left,
title={},
x grid style={darkgray176},
xticklabel style={/pgf/number format/fixed,
                  /pgf/number format/1000 sep=},
xlabel={Year},
xmin=1984.2, xmax=2023.8,
xtick style={color=black},
y grid style={darkgray176},
ylabel={Number of Papers},
ymin=-4104.9, ymax=86224.9,
ytick style={color=black}
]
\addplot [line width=2pt, steelblue31119180]
table {%
1990 2
1991 3
1992 1
1993 7
1994 251
1995 255
1996 236
1997 193
1998 334
1999 324
2000 512
2001 613
2002 702
2003 865
2004 1020
2005 1387
2006 1898
2007 2839
2008 3642
2009 4878
2010 7576
2011 9130
2012 12324
2013 14938
2014 16313
2015 18832
2016 23708
2017 30811
2018 41913
2019 55337
2020 71386
2021 77574
2022 82119
};
\addlegendentry{cs}
\addplot [line width=2pt, darkorange25512714]
table {%
1989 6
1990 24
1991 60
1992 358
1993 604
1994 947
1995 1199
1996 1417
1997 1850
1998 2696
1999 3338
2000 4091
2001 4347
2002 5639
2003 6641
2004 8344
2005 10023
2006 11990
2007 14241
2008 15512
2009 17574
2010 21175
2011 24159
2012 27237
2013 30218
2014 32102
2015 34744
2016 36404
2017 38468
2018 40003
2019 42952
2020 46099
2021 45163
2022 45415
};
\addlegendentry{math}
\addplot [line width=2pt, forestgreen4416044]
table {%
1986 1
1993 1
1994 45
1995 125
1996 237
1997 511
1998 832
1999 1043
2000 1599
2001 1654
2002 1699
2003 2120
2004 2674
2005 3362
2006 3914
2007 4416
2008 4858
2009 5691
2010 7428
2011 8688
2012 10390
2013 11196
2014 11666
2015 12442
2016 13723
2017 14571
2018 16885
2019 18969
2020 22064
2021 20804
2022 20262
};
\addlegendentry{physics}
\addplot [line width=2pt, crimson2143940]
table {%
1992 2
1993 4
1994 7
1995 9
1996 11
1997 20
1998 37
1999 70
2000 70
2001 113
2002 167
2003 469
2004 687
2005 686
2006 750
2007 945
2008 950
2009 1034
2010 1234
2011 1418
2012 1766
2013 2333
2014 2210
2015 2299
2016 2462
2017 2487
2018 2726
2019 2918
2020 4250
2021 3306
2022 3227
};
\addlegendentry{q-bio}
\addplot [line width=2pt, mediumpurple148103189]
table {%
1997 20
1998 42
1999 53
2000 77
2001 102
2002 114
2003 127
2004 146
2005 201
2006 259
2007 303
2008 324
2009 398
2010 526
2011 571
2012 628
2013 739
2014 819
2015 856
2016 916
2017 894
2018 1059
2019 1375
2020 1846
2021 1989
2022 1780
};
\addlegendentry{q-fin}
\addplot [line width=2pt, sienna1408675]
table {%
1994 1
1997 1
1998 1
1999 2
2000 9
2001 17
2002 26
2003 37
2004 162
2005 304
2006 510
2007 852
2008 1097
2009 1104
2010 1461
2011 1992
2012 3191
2013 3411
2014 3898
2015 4687
2016 5671
2017 7546
2018 12136
2019 16892
2020 17844
2021 10403
2022 10236
};
\addlegendentry{stat}
\end{axis}

\end{tikzpicture}}
    \caption{Number of Papers for Each Field (Up to 2023)}
    \label{fig: num-per-year-arxiv}
\end{figure}
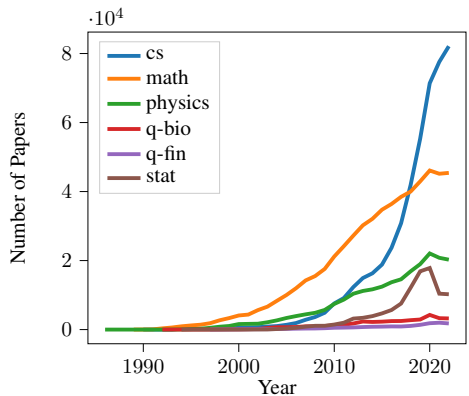



\subsection{Zero-shot} \label{ap: zero-shot-prompt}
\vspace{1em} 
\begin{minipage}{\columnwidth}
\begin{mdframed}[linewidth=1pt, roundcorner=10pt, skipabove=5pt, skipbelow=5pt, innermargin=5pt, outermargin=5pt]

\small

\noindent

\vspace{0.5em} 

You will be provided with the title and abstract of two research papers. Please determine which of the two articles is more novel. Follow these steps for evaluation.

\vspace{0.5em}

\textbf{Step 1: Identify the problem and solution that the research paper attempts to solve.}

\vspace{0.5em}

\textbf{Step 2: Determine how unique the solution is given the current research landscape in 2024.} Does the paper introduce a new idea, theory, or concept that has not been previously discussed in the literature?

\vspace{0.5em}

\textbf{Step 3: Determine how creative the solution is given the current research landscape in 2024.} Does it apply a known idea in a completely new context or in a way that has not been done before?

\vspace{0.5em}

\textbf{Step 4: Using the findings from Steps 1-3, determine which paper is more novel.}

\vspace{0.5em}

In your response, please only state which paper is more novel (e.g., 1 if Paper 1 is more novel; 2 if Paper 2 is more novel).

\vspace{1em} 
\textbf{User Prompt:}
\vspace{-5pt} 
\begin{itemize}
    \itemsep0pt
    \parskip0pt
    \item Paper 1 Title: [paper\_1\_title]
    \item Paper 1 Abstract: [paper\_1\_abstract]
    \item Paper 2 Title: [paper\_2\_title]
    \item Paper 2 Abstract: [paper\_2\_abstract]
\end{itemize}

\end{mdframed}
\end{minipage}

\vspace{1em}  
\subsection{Self-Reflection}\label{ap: prompt-self-reflection}
\begin{minipage}{\columnwidth}
\begin{mdframed}[linewidth=1pt, roundcorner=10pt, skipabove=5pt, skipbelow=5pt, innermargin=5pt, outermargin=5pt]

\small

\noindent

You are an advanced language model tasked with determining the novelty of research papers in 2024. Your goal is to evaluate and compare the novelty of two research papers based on their titles, abstracts, and any other given metadata.

\textbf{The order in which the papers are presented is random and should not influence your evaluation.}

\textbf{Step 1: Independent Evaluation}

Analyze each research paper's title and abstract \textbf{independently}. Treat each paper as if it is the only one under review at that moment.

Consider the following aspects for each paper:
\begin{itemize}
    \itemsep0pt
    \parskip0pt
    \item \textbf{Novelty of Methodology}: Are the methods used new and innovative?
    \item \textbf{Surprisingness of Findings}: Are the findings unexpected or counterintuitive?
    \item \textbf{Impact on Existing Knowledge}: How does the research challenge or expand current scientific understanding?
    \item \textbf{Potential for Future Research}: Does the paper open up new directions for research?
    \item \textbf{Relevance to 2024 Scientific Understanding}: How well does the paper align with or push the boundaries of current trends?
\end{itemize}

\vspace{-5pt}
\textbf{Step 2: Quantitative Assessment}
\vspace{-5pt}
\begin{itemize}
    \itemsep0pt
    \parskip0pt
    \item Assign a score from 1-10 to each research paper for its novelty, with 10 being the most novel. This score should be based solely on the content of the title and abstract.
    \item Provide a brief justification for the score, using specific quotes and context.
\end{itemize}

\vspace{-5pt}
\textbf{Step 3: Final Comparison}
\vspace{-5pt}
\begin{itemize}
    \itemsep0pt
    \parskip0pt
    \item After independently scoring each paper, compare the scores.
    \item Determine which paper exhibits greater novelty based on the higher score, and provide the identifier (X or Y) of the more novel paper.
\end{itemize}

\vspace{-5pt} 
\textbf{Important:} The order of presentation is random and should not influence your decision. Evaluate each paper strictly on its content and merit.

\vspace{1em} 
\textbf{User Prompt:}
\vspace{-5pt} 
\begin{itemize}
    \itemsep0pt
    \parskip0pt
    \item Paper X Title: [paper\_x\_title]
    \item Paper X Abstract: [paper\_x\_abstract]
    \item Paper Y Title: [paper\_y\_title]
    \item Paper Y Abstract: [paper\_y\_abstract]
\end{itemize}

\end{mdframed}
\end{minipage}

\vspace{1em}

\subsection{SchNovel}\label{ap: scinovel}

\begin{algorithm}[H]
\caption{Data Sampling Algorithm}\label{alg: sampling}
\small
\begin{algorithmic}
\State $Fields \gets [cs, math, physics, q\-bio, q\-fin, stat]$
\State $startYear \gets [2019, 2020, 2021, 2022, 2023]$
\State $yearGap \gets [2, 4, 6, 8, 10]$
\State $sampleNum \gets 100$
\State $N \gets 0$
\State $Dataset \gets []$
\For{\texttt{f in Fields}}
    \For{\texttt{s in startYear}}
        \For{\texttt{g in yearGap}}
            \While{$N \neq sampleNum$}
                \State $x \gets \texttt{paper published in s from f}$
                \State $y \gets \texttt{paper published in s-g from f}$ 
                \State $label \gets x$
                \State $Dataset \gets (f, g, s, x, y, label)$
                \State $N \gets N + 1$
            \EndWhile
            \State $N \gets 0$
        \EndFor
    \EndFor
\EndFor
\end{algorithmic}
\end{algorithm}

\subsection{LLM Discussion} \label{ap: LLM Discussion}
\vspace{1em} 
\begin{minipage}{\columnwidth}
\begin{mdframed}[linewidth=1pt, roundcorner=10pt, skipabove=5pt, skipbelow=5pt, innermargin=5pt, outermargin=5pt]

\small

\noindent

\vspace{0.5em} 

You are a [Role] with expertise across all areas of [Category]. You will be provided with the titles and abstracts of two research papers. Your task is to determine which of the two articles is more novel by evaluating their originality, contribution to the field, and potential impact. Focus on aspects such as new methodologies, unexplored problems, innovative solutions, and how the work advances the state of the art. Follow these steps for evaluation.

\vspace{0.5em}

\textbf{Step 1: Identify the problem and solution that the research paper attempts to solve.}

\vspace{0.5em}

\textbf{Step 2: Determine how unique the solution is given the current research landscape in 2024.} Does the paper introduce a new idea, theory, or concept that has not been previously discussed in the literature?

\vspace{0.5em}

\textbf{Step 3: Determine how creative the solution is given the current research landscape in 2024.} Does it apply a known idea in a completely new context or in a way that has not been done before?

\vspace{0.5em}

\textbf{Step 4: Using the findings from Steps 1-3, determine which paper is more novel.}

\vspace{0.5em}

Please limit your response to 150 tokens max. In your response please conclude with: "The more novel and impactful paper is [Paper X or Paper Y]

\vspace{1em} 
\textbf{User Prompt:}
\vspace{-5pt} 
\begin{itemize}
    \itemsep0pt
    \parskip0pt
    \item Paper X Title: [paper\_x\_title]
    \item Paper X Abstract: [paper\_x\_abstract]
    \item Paper Y Title: [paper\_y\_title]
    \item Paper Y Abstract: [paper\_y\_abstract]
    \item \textit{(Round 2 Discussion add on)} 
    [previous\_response]
    These are responses from other reviewers. Please revise your response if necessary... [other\_responses]
    \item \textit{(Round 3 Discussion add on)} These are responses from other reviewers. Please determine which paper is more novel... [other\_responses]
\end{itemize}

\end{mdframed}
\end{minipage}

\vspace{1em}  
\subsection{RAG-Novelty} \label{ap: prompt-rag-novelty}
\begin{minipage}{\columnwidth}
\begin{mdframed}[linewidth=1pt, roundcorner=10pt, skipabove=5pt, skipbelow=5pt, innermargin=5pt, outermargin=5pt]

\small

\noindent

You are an advanced language model tasked with determining the novelty of research papers in 2024. Your goal is to evaluate and compare the novelty of two research papers based on their titles and abstracts. 

\textbf{The order in which the papers are presented is random and should not influence your evaluation.}

\textbf{Step 1: Independent Evaluation}

Analyze each research paper’s title and abstract \textbf{independently}. Treat each paper as if it is the only one under review at that moment.

Retrieve similar abstracts from a vector database based on the provided abstracts.

\textbf{Contextual Date Analysis:} Average the published dates of the retrieved documents. Use this average date as additional context for your evaluation.
Consider that papers with an average date that is later or more recent in time are generally more novel.

Consider the following aspects for each paper:
\vspace{-5pt}
\begin{itemize}
    \itemsep0pt
    \parskip0pt
    \item \textbf{Novelty of Methodology:} Are the methods used new and innovative?
    \item \textbf{Surprisingness of Findings:} Are the findings unexpected or counterintuitive?
    \item \textbf{Impact on Existing Knowledge:} How does the research challenge or expand current scientific understanding?
    \item \textbf{Potential for Future Research:} Does the paper open up new directions for research?
    \item \textbf{Relevance to 2024 Scientific Understanding:} How well does the paper align with or push the boundaries of current trends?
\end{itemize}

\vspace{-5pt}
\textbf{Step 2: Quantitative Assessment}
\vspace{-5pt}
\begin{itemize}
    \itemsep0pt
    \parskip0pt
    \item Assign a score from 1-10 to each research paper for its novelty, with 10 being the most novel. This score should be based on the content of the title and abstract, as well as the contextual information from the average published dates.
    \item Provide a brief justification for the score, using specific quotes and context.
\end{itemize}

\vspace{-5pt}
\textbf{Step 3: Final Comparison}
\vspace{-5pt}
\begin{itemize}
    \itemsep0pt
    \parskip0pt
    \item After independently scoring each paper, compare the scores.
    \item Determine which paper exhibits greater novelty based on the higher score, and conclude with: "The more novel and impactful paper is [Paper X or Paper Y].
\end{itemize}

\vspace{-5pt}
\textbf{Important:} The order of presentation is random and should not influence your decision. Evaluate each paper strictly on its content and merit, incorporating the additional context from the vector database as described.

\vspace{1em} 
\textbf{User Prompt:}
\vspace{-5pt} 
\begin{itemize}
    \itemsep0pt
    \parskip0pt
    \item Paper X Average Cosine Similarity: [paper\_x\_avg\_cosine\_similarity]
    \item Paper X Average Contextual Date: [paper\_x\_avg\_contextual\_date]
    \item Paper Y Average Cosine Similarity: [paper\_y\_avg\_cosine\_similarity]
    \item Paper Y Average Contextual Date: [paper\_y\_avg\_contextual\_date]
    \item Paper X Title: [paper\_x\_title]
    \item Paper X Abstract: [paper\_x\_abstract]
    \item Paper Y Title: [paper\_y\_title]
    \item Paper Y Abstract: [paper\_y\_abstract]
\end{itemize}

\end{mdframed}
\end{minipage}

\end{document}